\documentclass[journal]{IEEEtran}
\usepackage{amsmath}
\usepackage[table,xcdraw]{xcolor}

\usepackage{times}
\usepackage{graphicx}
\usepackage{epstopdf}
\usepackage{epsfig}
\usepackage{amsmath}
\usepackage{amssymb}
\usepackage{soul}
\usepackage{url}
\usepackage[hidelinks]{hyperref}
\usepackage[utf8]{inputenc}
\usepackage[small]{caption}
\usepackage{graphicx}
\usepackage{amsmath}
\usepackage{amsthm}
\usepackage{booktabs}
\usepackage{algorithm}
\usepackage{algorithmic}
\usepackage[switch]{lineno}
\usepackage{multirow}
\usepackage{makecell}
\usepackage{color}
\usepackage{subfigure}
\usepackage{subfloat}
\usepackage{cleveref}

\ifCLASSINFOpdf
\else
\fi

\begin{document}


\title{Low-Light Enhancement Effect on Classification and Detection: An Empirical Study}

\author{Xu Wu, Zhihui Lai, Zhou Jie, Can Gao, Xianxu Hou, Ya-nan Zhang, Linlin Shen

\thanks{X. Wu, Z. Lai, Can Gao, Ya-nan Zhang and L. Shen are with the Computer Vision Institute, College of Computer Science and Software Engineering, Shenzhen University, Shenzhen 518060, China, Shenzhen Institute of Artificial Intelligence and Robotics for Society, Shenzhen 518060, China, and Guangdong Key Laboratory of Intelligent Information Processing, Shenzhen University, Shenzhen 518060, China(e-mail: csxunwu@gmail.com, lai\_zhi\_hui@163.com; zyn962464@gmail.com; llshen@szu.edu.cn).}
\thanks{X. Hou is with School of AI and Advanced Computing, Xi’an Jiaotong-Liverpool University, China (hxianxu@gmail.com).}
\thanks{J. Zhou is with the National Engineering Laboratory for Big Data System Computing Technology, Shenzhen University, and SZU Branch, Shenzhen Institute of Artificial Intelligence and Robotics for Society, Shenzhen, Guangdong 518060, China (e-mail: jie\_jpu@163.com).}
}

\markboth{Journal of \LaTeX\ Class Files,~Vol.~14, No.~8, August~2015}%
{Shell \MakeLowercase{\textit{et al.}}: Bare Demo of IEEEtran.cls for IEEE Journals}

\maketitle

\begin{abstract}
Low-light images are commonly encountered in real-world scenarios, and numerous low-light image enhancement (LLIE) methods have been proposed to improve the visibility of these images. The primary goal of LLIE is to generate clearer images that are more visually pleasing to humans. 
However, the impact of LLIE methods in high-level vision tasks, such as image classification and object detection, which rely on high-quality image datasets, is not well {explored}. 
To explore the impact, we comprehensively evaluate LLIE methods on these high-level vision tasks by utilizing an empirical investigation comprising image classification and object detection experiments. The evaluation reveals a dichotomy: {\textit{While Low-Light Image Enhancement (LLIE) methods enhance human visual interpretation, their effect on computer vision tasks is inconsistent and can sometimes be harmful. }} Our findings suggest a disconnect between image enhancement for human visual perception and for machine analysis, indicating a need for LLIE methods tailored to support high-level vision tasks effectively. This insight is crucial for the development of LLIE techniques that align with the needs of both human and machine vision.
\end{abstract}


%
\IEEEpeerreviewmaketitle

\section{Introduction}
\label{sec:intro}
Low-light conditions refer to situations with insufficient light to clearly see an image or scene. This can occur during nighttime, or when objects are partially or fully in shadow. As a result, a significant proportion of images captured in these conditions, which are called low-light images, are degraded in terms of visibility and clarity.  The prevalence of low-light images in real-world scenarios has motivated the development of techniques for improving their visibility and clarity, known as low-light image enhancement (LLIE) methods.

\begin{figure}[htbp]
\centering
\includegraphics[width=8.25 cm]{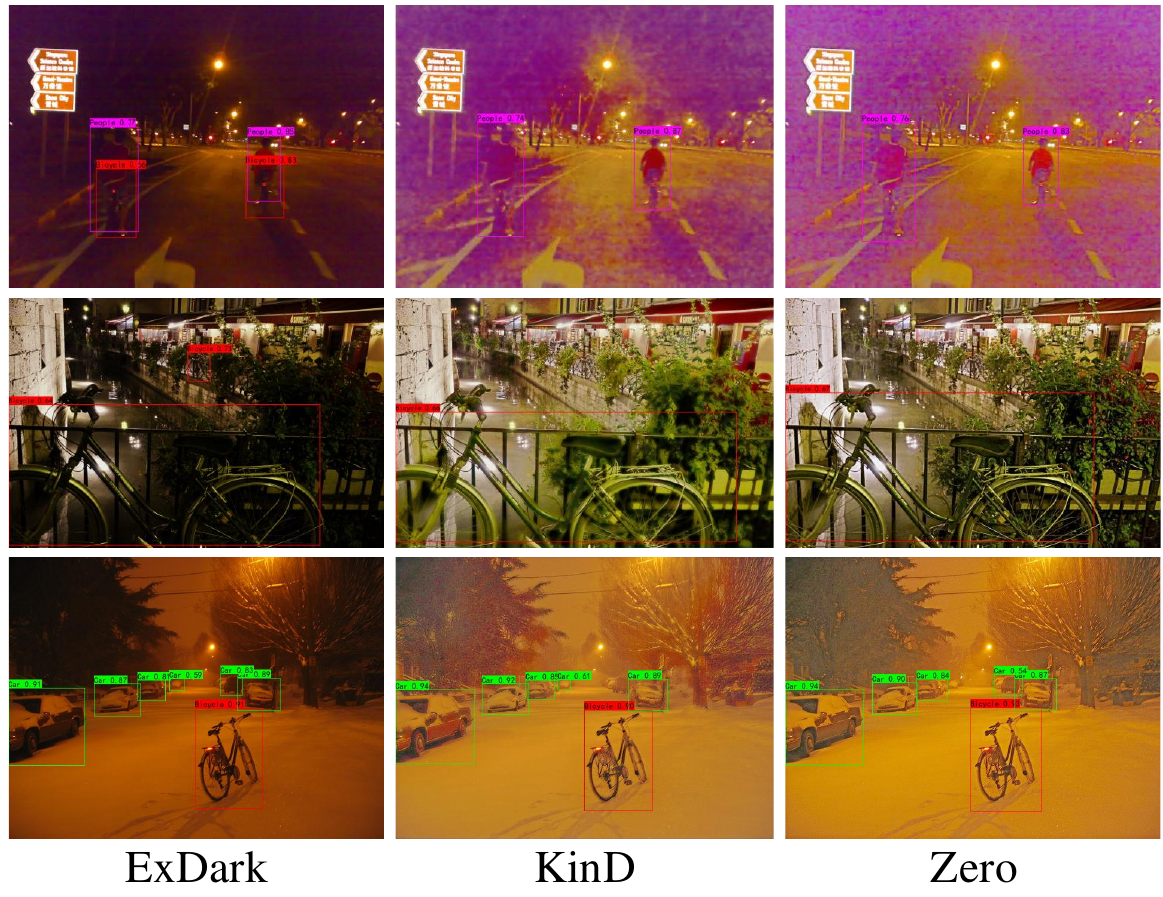}
\caption{Object detection examples of YOLOv7 \cite{yolov7} on ExDark \cite{Exdark} and light-enhanced ExDark (using KinD \cite{KinD} and Zero \cite{Zero}) dataset.}
\vspace{-1.2 em}
\label{figs1}
\end{figure}
LLIE methods can be divided into two {categories}: traditional and deep learning-based. Traditional methods include techniques such as histogram equalization (HE) and retinex theory \cite{retinex}, which are commonly used to improve the visibility and perception of low-light images. HE homogenizes the pixel value intensity of an image to adjust its brightness, while the retinex theory separates an image into its reflection and illumination components for processing individually. 
Traditional methods, while effective in specific scenarios, are constrained by their dependence on predefined priors and assumptions about low-light image characteristics. In contrast, deep learning approaches excel by autonomously learning features and patterns from data, yielding more robust and effective outcomes in low-light conditions.
Deep learning-based LLIE methods have gained significant attention in recent years due to their potential for improving the clarity and visibility of low-light images. These methods can take various forms, including different network architectures \cite{LRCR}, loss functions \cite{Zero}, supervised \cite{KinD}, and unsupervised \cite{EnGAN} learning approaches. 
Furthermore, peak signal-to-noise ratio (PSNR) and structural similarity (SSIM) are prevalent metrics for evaluating image quality in these methods. These metrics are specifically designed to assess the fidelity of enhanced images to the ground truth and primarily concentrate on the subjective quality of the enhanced image.
Consequently, these methods prioritize enhancing visual appeal for human observers rather than boosting the performance of high-level tasks like image classification and object detection in low-light scenarios.

In general, the impact of LLIE methods on high-level tasks can be analyzed from two perspectives:
\begin{itemize}
    \item LLIE methods enhance the quality of low-light images by augmenting illumination and restoring color information. This process reveals details previously obscured in darkness, thereby improving image clarity. Consequently, LLIE not only improves visual quality for human viewers but also potentially enhances the performance of high-level tasks.
    
    \item LLIE methods aim to enhance visibility and clarity in low-light images without adding new information or modifying the semantics of the images. From this point, these methods cannot inherently boost the performance of high-level algorithms that depend on extracting feature or semantic information from input images. As illustrated in Fig. \ref{figs1}, the performance of such algorithms is not significantly degraded as they can effectively extract sufficient information from low-light images.
\end{itemize}

To explore the effect of LLIE methods on the performance of high-level tasks under low-light conditions, we conduct a series of carefully designed experiments. 
We first construct a low-light image classification dataset to accurately assess the performance of image classification algorithms in low-light environments. The dataset contains a total of 3,853 images across 19 categories, with an average of 200 images per category. These images are collected from real-world low-light scenes. Simultaneously, we propose a method for synthesizing low-light images from high-quality images to obtain \textit{normal-low} light image pairs. This allowed us to quantify the impact of low-light conditions on the performance of image classification algorithms and to analyze the effectiveness of low-light image enhancement methods in terms of image quality improvement and performance improvement of image classification and object detection algorithms.
Moreover, for the LLIE task, we select 3 traditional and 9 deep learning-based methods to enhance low-light images. 
For high-level tasks, we exploit image classification and object detection tasks as representative high-level tasks, as they are commonly used in various applications. These tasks have been widely studied and are trained using datasets such as Caltech-256 \cite{caltech256}, ImageNet \cite{ImageNet}, and COCO \cite{COCO}, which mainly contain images with normal illumination. 

To sum up, we conduct extensive experiments to explore the effectiveness of LLIE on high-level task in low-light conditions. We also provide experimental insights and discussion for further research on LLIE methods, {aiming} to improve high-level task performance in low-light environments.
Overall, our contributions are listed as follows:
\begin{itemize}
    \item We concentrate on investigating the impact of the LLIE methods for high-level tasks in low-light conditions through the comprehensive analysis of LLIE, image classification, and object detection.

    \item A new low-light image dataset, comprising 19 categories and 3,853 low-light images, is constructed for image classification tasks in low-light conditions.

    \item {Our experimental results suggest that LLIE methods do not notably enhance the performance in high-level tasks under low-light conditions. In fact, they may even reduce the effectiveness of these tasks, indicating a counterproductive effect in certain applications.}
\end{itemize}

\section{Related Works}
\subsection{Low-Light Image Enhancement Methods}
LLIE aims to improve the visibility of images captured in low-light conditions by increasing the overall brightness, reducing noise, and increasing contrast \cite{LRCR}\cite{LACR}. 
Dong \cite{Dong} performs LLIE by using the dehazing algorithm to process the reverse image of the low-light image.
JED \cite{JED} concurrently addresses both low-illumination enhancement and image denoising, effectively suppressing noise while achieving illumination enhancement.
LIME \cite{LIME} estimates the illumination of the input image by finding the maximum value of each channel. In addition, a substructure is designed to refine the initial illumination map.
EnGAN \cite{EnGAN} extracts image features to regularize unsupervised training and designs a global-local discriminator structure for LLIE.
KinD \cite{KinD} lighting low-light image enhancement by incorporating two main components: an illumination adjustment network and a content restoration network. 
RetinexNet \cite{RetinexNet} decomposes input image into illumination and reflection components using a decomposition sub-network and then performs image enhancement using an enhancement sub-network. 
Zero \cite{Zero} converts low-light enhancement tasks into estimates image special curve tasks. 
UTV \cite{UTVNet} presents an adaptive unfolding total variational network for restoring low-light images in the RGB color space. 
SCI \cite{SCI} enhances the illumination of an input image by a self-calibrated illumination learning framework. 
The PSNR is used to guide enhancement in SNR \cite{SNR} and the image structure information is introduced to improve output quality in SMG \cite{SMG}. PairLIE \cite{PairLIE} proposes an unsupervised strategy to perform LLIE. To effectively use high-quality images prior, CodeEnhance \cite{wu2024codeenhance} introduces the codebook \cite{VQGAN}\cite{VQVAE} into LLIE to achieve outstanding performance.

The objective of the methods mentioned above is to enhance the visual quality of low-light images, and their performance is evaluated using image metrics that are based on human visual perception. Therefore, their effectiveness in improving the performance of high-level tasks under low-light conditions is worth exploring.

\subsection{Image Classification and Object Detection}
Image classification and object detection are fundamental tasks in computer vision and have been well-studied. Image classification aims to automatically assign a label or a class to images. Various architectures of deep convolution neural networks (CNNs) and Transformer \cite{transformer} are commonly used in image classification, such as VGG \cite{vgg}, ResNet \cite{resnet}, DenseNet \cite{densenet}, and Vision Transformer (ViT) \cite{vit}. Object detection methods mainly include two categories: two-stage and one-stage. Two-stage algorithms \cite{rcnn} generate candidate regions and then classify them. However, one-stage algorithms like YOLOv7 \cite{yolov7}, SSD \cite{ssd}, and RetinaNet \cite{retinanet} directly predict the objects' category probability and position coordinates. 
RetinaNet proposed Focal Loss, designed to reduce overfitting on easy samples and increase focus on hard samples.
EfficientDet \cite{efficientdet} is based on the idea of feature pyramid networks, which are designed to extract high-level features from input images and use them to detect objects at multiple scales. 
YOLO (You Only Look Once) \cite{yolo} is a real-time object detection algorithm that can detect and classify objects in an image or video in a single pass. 


In this paper, to explore the impact of LLIE on performing high-level tasks, we implement CNN-based models for image classification by using VGG19\cite{vgg}, ResNet50\cite{resnet}, DenseNet121\cite{densenet}, and ViT\cite{vit}. VGG19, ResNet50, and DenseNet121 are well-known networks in image classification tasks. For object detection, we select three representative methods for evaluation: RetinaNet\cite{retinanet}, EfficientDet\cite{efficientdet}, and YOLOv7\cite{yolov7}.



\begin{figure}[]
\centering
\includegraphics[width=8.25 cm]{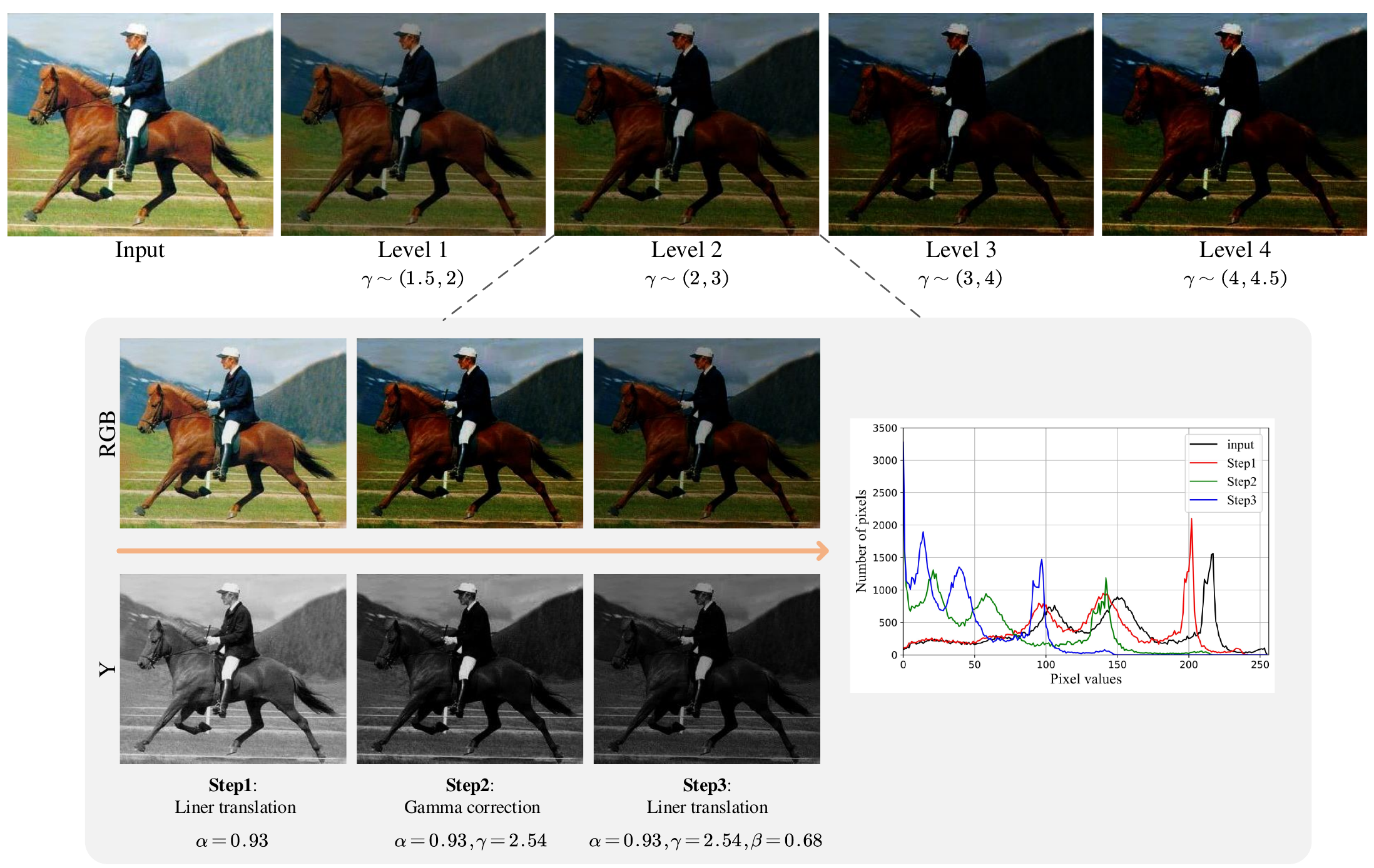}
\caption{Visual multi-level illumination of our low-light synthesis. }
\label{fig2}
\end{figure}

\section{Datasets}
In this section, we investigate the details of the low-light image datasets, LLIE methods, high-level task frameworks, and evaluation metrics used in this study.
We utilize 3 datasets for classification and 1 dataset for object detection, which are detailed as follows:

\subsection{Datasets for Classification}

To explore the relationship between LLIE and image classification tasks, we require a large dataset of low-light images for both training and testing. Considering current image classification datasets, like ImageNet \cite{ImageNet} and Caltech-256 \cite{caltech256}, primarily consist of clear images, we collect two low-light image datasets for our study.

\textbf{Caltech256.} The Caltech-256 \cite{caltech256} dataset has been widely used for training and evaluating image classification methods, which consists of 30,607 images from 257 classes (256 object classes and a clutter class). In our experiment, we select $70\%, 10\%$, and $20\%$ images randomly from each class as training, validation, and testing images, respectively. This way, we have a training set of 21,424 images, a validation set of 3,060 images, and a testing set of 6,123 images.

\textbf{Caltech256-LL.} We create a synthetic low-light image dataset called Caltech256-LL based on the Caltech-256 dataset. To synthesize this dataset, we use gamma correction and linear translation techniques to adjust the illumination of the images:
\begin{equation}
    I_{ll} = \beta \times (\alpha \times I_{cl})^{\gamma}
\end{equation}
where $\beta \sim \mathcal{U}(0.5, 1.0)$ and $\alpha \sim \mathcal{U}(0.9, 1.0)$ are used to perform the linear translation. $(\cdot)^{\gamma}$ means the gamma correction. $I_{cl}$ and $I_{ll}$ represents clear and low-light image.
In our observations, the illumination intensity can be divided into multiple levels depending on the time of day and the light source. Therefore, we set four different illumination levels for generating low-light images. Specifically, we set $\gamma \in \{\gamma_1, \gamma_2, \gamma_3, \gamma_4\}$, $\gamma_1 \sim [1.5, 2], \gamma_2 \sim [2, 3], \gamma_3 \sim [3, 4]$, and $\gamma_4 \sim [4, 4.5]$. 
In our experiment, we select $70\%, 10\%$, and $20\%$ images randomly from each class as training, validation, and testing images, respectively. This way, we have a training set of 21,424 images, a validation set of 3,060 images, and a testing set of 6,123 images.

\textbf{Dark Class dataset.} By using gamma correction and linear translation, we can synthetic realistic low-light images based on high-quality images. However, the distribution of low-light images in the real world is complex, and the synthesis process cannot simulate the distribution perfectly. To solve this drawback and evaluate the image classification performance of models in real-world low-light scenes, we collect a new low-light image dataset (called Dark Class) from the Internet and mobile phones.
The Dark Class contains 3,853 low-light images from multi-level illumination and 19 categories (airline (216), arch bridge (201), modern arch bridge (161), suspension bridge (253), bike (213), mountain bike (203), building (213), bus (234), car (241), coach (168), cruise (190), motorcycle (207), people (197), skyscraper (201), tower (195), traffic light (197), transport plane (144), truck (216), yacht (203)). We set the training, validation, and testing parts as $7:1:2$. Therefore, we have 2,688 training images, 378 validation images, and 787 testing images.
We believe this dataset can help researchers explore the characteristics of low-light image features and design better networks for learning low-light image information.

\begin{figure}[]
\centering
\includegraphics[width=8.5 cm]{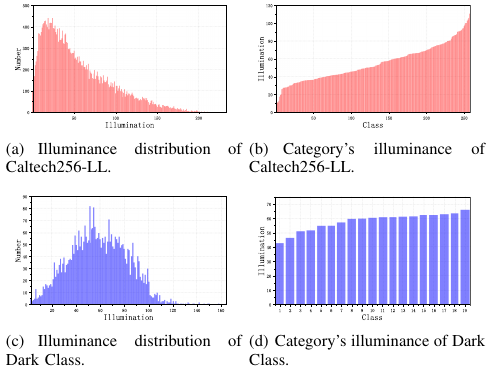}
\caption{Illuminance information's statistics of the proposed Caltech256-LL and Dark Class dataset.}
\vspace{-0.5em}
\label{fig4}
\end{figure}

\begin{figure}[]
\centering
\includegraphics[width=8.25 cm]{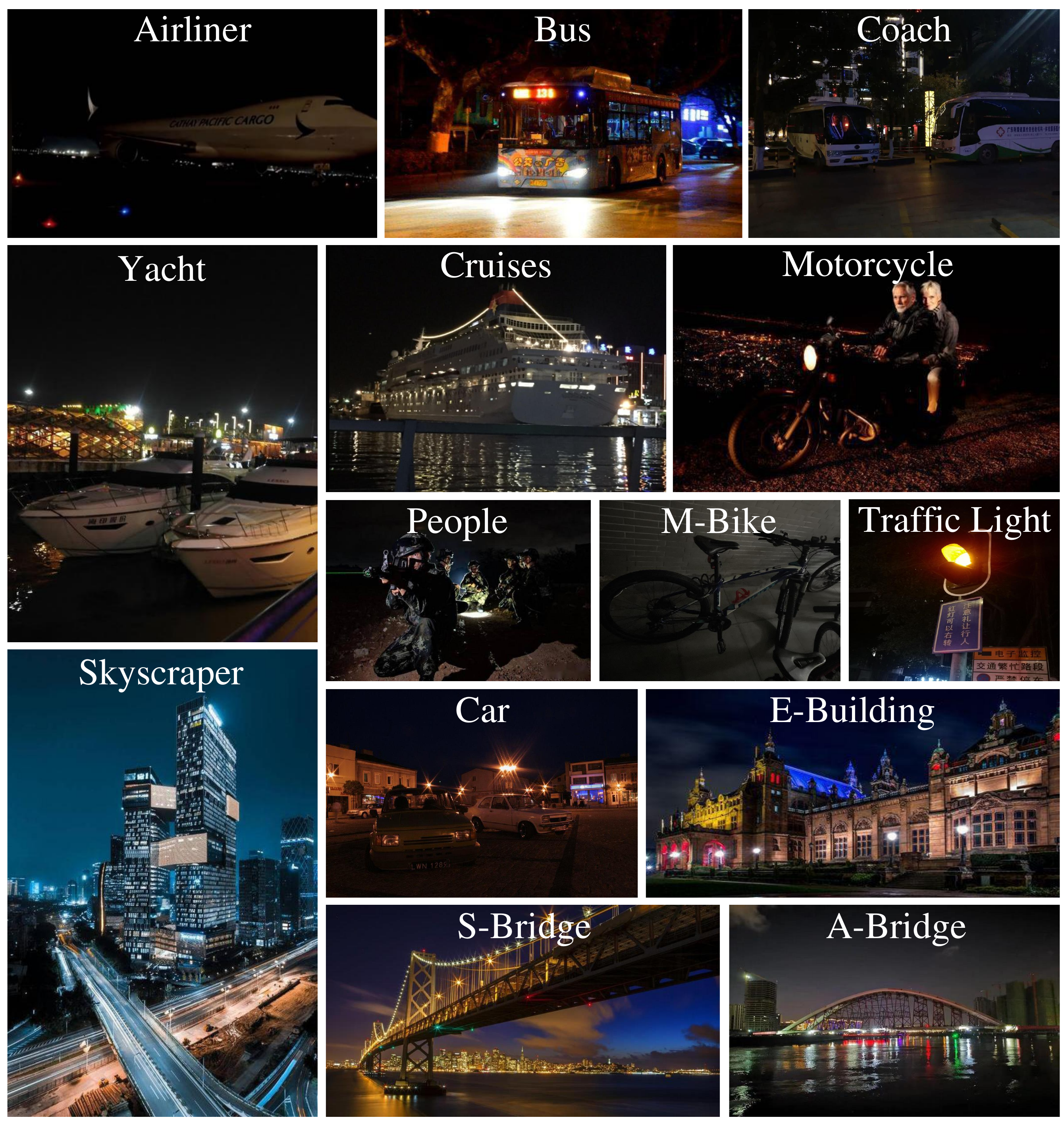}
\caption{Example images from the proposed Dark Class dataset. '\textit{Bike-M}' means mountain bike.}
\label{fig3}
\end{figure}

As illustrated in Fig. \ref{fig4} (c)-(d), the illuminance distribution of our Dark Class follows a normal distribution. The majority of images exhibit an average pixel value within the range of 40-80, and the average illuminance distribution across all categories is homogeneous. This broad range of illuminance levels is beneficial for the network to learn the illuminance distribution and to produce a low-light image classifier that is suitable for real-world scenarios. 
Some examples of the Dark Class dataset are shown in Fig. \ref{fig3}.




\subsection{Dataset for Object Detection}
\textbf{ExDark} \cite{Exdark} is a low-light image dataset for object detection. It contains 7,363 images from multi-level illumination and 12 object classes that are bicycle (652), boat (679), bottle (547), bus (527), car (638), cat (735), chair (648), cup (519), dog (801), motorbike (503), people (609), and table (505). Furthermore, we partitioned the dataset according to the strategy outlined in the ExDark, resulting in the obtaining of the training (3,000), validation (1,800), and test (2,563) sets.

\section{Experiments}
We conduct 114 experiments in this section. Specifically, 9 different LLIE methods are applied to 3 datasets as preprocessing, and 4 networks are used for image classification, while 3 models are used for object detection.  

\begin{table*}[]
\centering
\footnotesize
\caption{Quantitative comparisons of low-light image enhancement methods.}
\begin{tabular}{|c|c|c|c|c|c|c|c|c|}
\hline
Methods & \multicolumn{4}{c|}{Caltech256}                                          & \multicolumn{2}{c|}{Dark\_Class}    & \multicolumn{2}{c|}{ExDark}         \\ \hline 
Metrics                        & PSNR $\uparrow$             & SSIM  $\uparrow$          & BRISQUE $\downarrow$         & NIQE  $\downarrow$          & BRISQUE  $\downarrow$        & NIQE  $\downarrow$          & BRISQUE  $\downarrow$        & NIQE  $\downarrow$          \\ \hline
Dong                    & 15.3187          & 0.6436          & 34.3821          & 8.4101          & 35.0674          & 4.4369          & 34.4096          & 4.3843          \\
EnGAN                   & {\color{red}18.4882} & {\color{blue}0.7674} & 33.4950          & 8.9185          & 34.6768          & 4.3153          & 31.9125          & {\color{red}4.1475} \\
JED                     & 15.2626          & 0.6883          & 40.6831          & 9.9331          & 38.7245          & 4.8961          & 37.4563          & 4.8161          \\
KinD                    & 17.9733          & 0.7383        & 31.2903          & 8.6736          & {\color{blue}29.1203}          & {\color{red}4.0019} & {\color{red}29.2759}          & 4.3479          \\
LIME                    & 16.9034          & 0.6891          & 36.6941          & 8.7061          & 37.5604          & 4.2909          & 35.6326          & 4.1565          \\
RetinexNet              & 14.5004          & 0.6684          & 39.4280          & 8.8908          & 33.8800            & {\color{blue}4.1226}          & 33.2249          & 4.3232          \\
SCI                     & 17.1763          & 0.6781          & 36.4579          & 8.7838          & 34.6191          & 4.2689          & {\color{blue}29.7617} & 4.1813          \\
UTV                     & 13.7443          & 0.6725          & 28.0424 & 7.9671 & 32.3023          & 4.4353          & 31.3003          & 4.3133          \\
Zero                    & 15.8633          & 0.7246          & 34.1067          & 8.2131          & 37.4271          & 4.2931          & 34.4825          & {\color{blue}4.1478}          \\ 

SNR    &     14.3855      &    0.6281       &   {\color{blue} 26.1334}       & {\color{red}  7.6193  }      &   33.0499        &  4.8509          &       37.8610    &   5.7829        \\ 
PairLIE    &    {\color{blue}18.1842}       &   {\color{red}0.7741}        &    34.8982       &  8.3134         &    37.8083       &   4.4305        & 37.3771          &     4.3755          \\ 
SMG   & 15.8621 &	0.7025	&{\color{red}21.1633}&	{\color{blue}7.6407}&	{\color{red}27.8184}	&4.7739&	33.1974	& 5.1599       \\  

\hline
\end{tabular}
\label{table_LLIE}
\end{table*}

\subsection{Experiment Setup}
In this study, we utilize several classic networks for image classification and object detection tasks. Specifically, VGG19, ResNet50, DenseNet121, and ViT networks are used for image classification. RetinaNet, EfficientDet, and YOLOv7 are used for object detection. The backbone networks of all models are initialized with pre-trained weights on ImageNet and fine-tuned using low-light image data for our specific tasks. For image classification, we utilize the SGD optimizer with an initial learning rate of 0.001, a batch size of 64, and an input image size of $224 \times224$. A total of 100 training epochs are conducted. For object detection, we use the Adam optimizer with an initial learning rate of 0.001, batch size of 8, and conduct 200 training epochs. All experiments are implemented using PyTorch.

\subsection{Evaluation Metrics}
In our experiments, we will evaluate the performance of algorithms for LLIE, image classification, and object detection using various metrics. PSNR and SSIM\cite{ssim} are full-reference indicators commonly used to measure image quality. PSNR measures the difference between two images on a pixel-by-pixel basis, while SSIM measures the distance between two images based on contrast, brightness, and structure, taking into account human visual perception. PSNR can be formulated as follow:

\begin{equation}
    PSNR = 10 {\rm log}_{10}(\frac{(2^n - 1)^2}{{\rm MSE}(x,y)})
\end{equation}
where $x$ and $y$ mean reference image and enhanced image, $n$ is the number of binary bits per pixel. ${\rm MSE}(\cdot)$ represents mean square error.

No-reference image quality evaluation indicators, such as Natural Image Quality Evaluator (NIQE)\cite{niqe} and Blind/Referenceless Image Spatial Quality Evaluator (BRISQUE)\cite{brisque}, are commonly used when there is no reference image. NIQE is a statistical model for measuring the quality of images, based on a set of features derived from a collection of natural, undistorted images known as the Nature Scene Static (NSS) model.
Let $\mu_1, \mu_2$ and $\sigma_1, \sigma_2$ represent the mean vectors and covariance matrices of the Multivariate Gaussian Distribution (MVG) model and the input image's MVG model, NIQE is formulated as follows:

\begin{equation}
    {\rm D}(\mu_1, \mu_2, \sigma_1, \sigma_2) = \sqrt{(\mu_1 - \mu_2)^T (\frac{\sigma_1 + \sigma_2}{2})^{-1} (\mu_1 - \mu_2)}
\end{equation}

BRISQUE uses NSS-based wavelet coefficient models to quantify the loss of image quality due to distortions. It uses a distortion identification and distortion-specific quality assessment framework to measure the quality of an image. Overall, a better image quality means a higher PSNR and SSIM, and a lower BRISQUE and NIQE.

Overall accuracy (OA) and mean average precision (mAP) are commonly used metrics for evaluating image classification and object detection performance, respectively. The OA can be formulated as follow:

\begin{equation}
    {\rm A}(R, N) = \frac{R}{N}
\end{equation}
where $R$ is the number of correctly classified test images and $N$ means the total number of test images. In this study, we will investigate the relationship between PSNR, SSIM, BRISQUE, NIQE, OA, and mAP for images that have been preprocessed using LLIE methods.

\subsection{Evaluation of LLIE methods}
We evaluate 12 LLIE methods using Caltech256-LL, Dark Class, and ExDark datasets. The PSNR, SSIM, Natural Image Quality Evaluator (NIQE) \cite{niqe}, and Blind/Referenceless Image Spatial Quality Evaluator (BRISQUE) \cite{brisque} are used to evaluate the enhancement performance of LLIE methods.
As shown in Table \ref{table_LLIE}, 
when the performance of the method is measured using the human perception-based image quality evaluation metrics, we find that the EnGAN, KinD, UTV, and Zero methods are superior.
In addition, we train the image classification models on Caltech256 and test them on Caltech256, Caltech256-LL, and light-preprocessed Caltech256-LL. Table \ref{table_LLIE} shows that the models perform best on Caltech256, followed by light-preprocessed Caltech256-LL and then Caltech256-LL. Currently, the UTV, RetinexNet, LIME, and Zero are superior. 
Indeed, these methods are fundamentally designed to enhance the visual perception of low-light images, enabling them to achieve commendable scores. Following this, our study explores how their advancements in illumination enhancement and color restoration for low-light images influence the performance of high-level tasks. 






\begin{figure}[]
\centering
\includegraphics[width=8.5 cm]{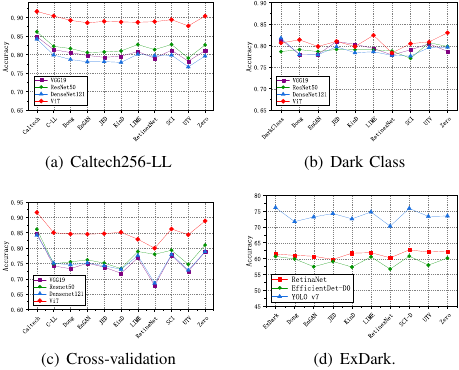}
\caption{(a) and (b) represent quantitative comparisons of image classification on the Caltech256-LL and Dark Class datasets. (c) denotes quantitative comparisons of the image classification trained on the Caltech and tested on the Caltech-LL dataset. d) represent quantitative comparisons of object detection on the ExDark dataset. }
\vspace{-3mm}
\label{fig-line}
\end{figure}



\begin{figure}[]
\centering
\includegraphics[width=8.5 cm]{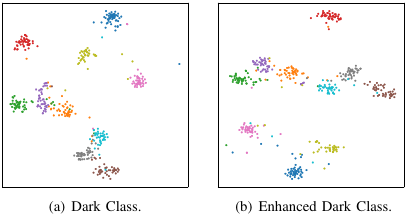}
\caption{Visual results of pixel features of ResNet50 trained on Dark Class, and Dark Class enhanced by UTV. The dimensionality reduction method is the t-SNE algorithm.}
\vspace{-4mm}
\label{pixel-feature}
\end{figure}

\subsection{Effect of Low-Light on Classification and Object Detection}
We train the models of image classification and object detection mentioned above on Caltech256 and test them on Caltech256-LL. As shown in Tables \ref{table_Caltech}-\ref{table_Dark_Ex} and Fig.\ref{fig-line} (c), the performance of image classification methods is significantly decreased when tested on low-light images. Specifically, the CNN classifier does worse by about 0.1, and the Transformer classifier does worse by about 0.06. The observed performance reduction can be attributed to the fundamental differences between high-quality and low-light images. Compared to high-quality images, low-light images exhibit reduced brightness and increased noise, leading to less pronounced image features and poorer feature quality. Consequently, classification and detection networks, which are trained on high-quality images, suffer to extract features from low-light images effectively. This mismatch results in a notable decline in their performance in low-light conditions.

\begin{table*}[]
\centering
\caption{The top table is the Overall Accuracy (OA, \%) comparison of image classification. All classifiers are trained on the Caltech256 \cite{caltech256} and tested on the Caltech256 and Caltech-LL datasets. The bottom table compares the overall accuracy for image classification on the Caltech256 dataset and the proposed Caltech256-LL dataset. 'NULL' means no preprocessing for dataset.}
\resizebox{\textwidth}{!}{
\begin{tabular}{ccccccccccccccc}
\hline
\multicolumn{1}{|c|}{Train   Set}            & \multicolumn{14}{c|}{Caltech256}                                                                                                                                                                                                                                                                                                                                                                                                      \\ \hline
\multicolumn{1}{|c|}{Test Set (TeS)}         & \multicolumn{1}{c|}{\cellcolor[HTML]{EFEFEF}Caltech256} & \multicolumn{13}{c|}{Caltech256-LL}                                                                                                                                                                                                                                                                                                                                                                 \\ \hline
\multicolumn{1}{|c|}{Preprocess for   (TeS)} & \multicolumn{1}{c|}{\cellcolor[HTML]{EFEFEF}NULL}       & \multicolumn{1}{c|}{NULL}   & \multicolumn{1}{c|}{Dong}   & \multicolumn{1}{c|}{EnGAN}  & \multicolumn{1}{c|}{JED}    & \multicolumn{1}{c|}{KinD}   & \multicolumn{1}{c|}{LIME}   & \multicolumn{1}{c|}{RetinexNet} & \multicolumn{1}{c|}{SCI}    & \multicolumn{1}{c|}{UTV}     & \multicolumn{1}{c|}{Zero}   & \multicolumn{1}{c|}{SNR} & \multicolumn{1}{c|}{PairLIE} & \multicolumn{1}{c|}{SMG} \\ \hline

\multicolumn{1}{|c|}{VGG}                    & \multicolumn{1}{c|}{\cellcolor[HTML]{EFEFEF}0.8481}     & \multicolumn{1}{c|}{0.7415} & \multicolumn{1}{c|}{0.7548} & \multicolumn{1}{c|}{0.7613} & \multicolumn{1}{c|}{0.7511} & \multicolumn{1}{c|}{0.7331} & \multicolumn{1}{c|}{0.7887} & \multicolumn{1}{c|}{{\color{blue}0.7928}}     & \multicolumn{1}{c|}{0.7471} & \multicolumn{1}{c|}{{\color{red}0.8098}} & \multicolumn{1}{c|}{0.7797} & \multicolumn{1}{c|}{0.7351}    & \multicolumn{1}{c|}{0.7378}        & \multicolumn{1}{c|}{0.7728}    \\ 

\multicolumn{1}{|c|}{Resnet}                 & \multicolumn{1}{c|}{\cellcolor[HTML]{EFEFEF}0.8614}     & \multicolumn{1}{c|}{0.7484} & \multicolumn{1}{c|}{0.7327} & \multicolumn{1}{c|}{0.7507} & \multicolumn{1}{c|}{0.7378} & \multicolumn{1}{c|}{0.7177} & \multicolumn{1}{c|}{0.7681} & \multicolumn{1}{c|}{0.7768}     & \multicolumn{1}{c|}{0.7237} & \multicolumn{1}{c|}{{\color{red}0.7879}}  & \multicolumn{1}{c|}{0.6777} & \multicolumn{1}{c|}{0.7512}    & \multicolumn{1}{c|}{0.7438}        & \multicolumn{1}{c|}{{\color{blue}0.7805}}    \\ 

\multicolumn{1}{|c|}{DenseNet}               & \multicolumn{1}{c|}{\cellcolor[HTML]{EFEFEF}0.8418}     & \multicolumn{1}{c|}{0.7507} & \multicolumn{1}{c|}{0.7452} & \multicolumn{1}{c|}{0.7515} & \multicolumn{1}{c|}{0.7422} & \multicolumn{1}{c|}{0.7307} & \multicolumn{1}{c|}{0.7754} & \multicolumn{1}{c|}{{\color{blue}0.7788}}     & \multicolumn{1}{c|}{0.7281} & \multicolumn{1}{c|}{{\color{red}0.7881}}  & \multicolumn{1}{c|}{0.6851} & \multicolumn{1}{c|}{0.7372}    & \multicolumn{1}{c|}{0.7365}        & \multicolumn{1}{c|}{0.7697}    \\ 

\multicolumn{1}{|c|}{ViT}                    & \multicolumn{1}{c|}{\cellcolor[HTML]{EFEFEF}0.9163}     & \multicolumn{1}{c|}{0.8501} & \multicolumn{1}{c|}{0.8459} & \multicolumn{1}{c|}{0.8454} & \multicolumn{1}{c|}{0.8469} & \multicolumn{1}{c|}{0.8514} & \multicolumn{1}{c|}{0.8287} & \multicolumn{1}{c|}{0.7995}     & \multicolumn{1}{c|}{0.8622} & \multicolumn{1}{c|}{0.8434}  & \multicolumn{1}{c|}{{\color{red}0.8883}} & \multicolumn{1}{c|}{0.8508}    & \multicolumn{1}{c|}{0.8633}        & \multicolumn{1}{c|}{{\color{blue}0.8757}}    \\ \hline \hline

\multicolumn{1}{|c|}{Dataset}                & \multicolumn{1}{c|}{\cellcolor[HTML]{EFEFEF}Caltech256} & \multicolumn{13}{c|}{Caltech256-LL}                                                                                                                                                                                                                                                                                                                                                                 \\ \hline
\multicolumn{1}{|c|}{Preprocess}             & \multicolumn{1}{c|}{\cellcolor[HTML]{EFEFEF}NULL}       & \multicolumn{1}{c|}{NULL}   & \multicolumn{1}{c|}{Dong}   & \multicolumn{1}{c|}{EnGAN}  & \multicolumn{1}{c|}{JED}    & \multicolumn{1}{c|}{KinD}   & \multicolumn{1}{c|}{LIME}   & \multicolumn{1}{c|}{RetinexNet} & \multicolumn{1}{c|}{SCI\_D} & \multicolumn{1}{c|}{UTV}     & \multicolumn{1}{c|}{Zero}   & \multicolumn{1}{c|}{SNR} & \multicolumn{1}{c|}{PairLIE} & \multicolumn{1}{c|}{SMG} \\ \hline

\multicolumn{1}{|c|}{VGG}                  & \multicolumn{1}{c|}{\cellcolor[HTML]{EFEFEF}0.8481}     & \multicolumn{1}{c|}{{\color{red}0.8129}} & \multicolumn{1}{c|}{0.8053} & \multicolumn{1}{c|}{0.7974} & \multicolumn{1}{c|}{0.7925} & \multicolumn{1}{c|}{0.7955} & \multicolumn{1}{c|}{0.8081} & \multicolumn{1}{c|}{0.7899}     & \multicolumn{1}{c|}{{\color{blue}0.8116}} & \multicolumn{1}{c|}{0.7818}  & \multicolumn{1}{c|}{0.8107} & \multicolumn{1}{c|}{0.7913}    & \multicolumn{1}{c|}{0.7971}        & \multicolumn{1}{c|}{0.8102}    \\ 

\multicolumn{1}{|c|}{ResNet}               & \multicolumn{1}{c|}{\cellcolor[HTML]{EFEFEF}0.8614}     & \multicolumn{1}{c|}{0.8221} & \multicolumn{1}{c|}{0.8163} & \multicolumn{1}{c|}{0.8053} & \multicolumn{1}{c|}{0.8071} & \multicolumn{1}{c|}{0.8097} & \multicolumn{1}{c|}{{\color{blue}0.8269}} & \multicolumn{1}{c|}{0.8132}     & \multicolumn{1}{c|}{{\color{red}0.8272}} & \multicolumn{1}{c|}{0.7905}  & \multicolumn{1}{c|}{0.8261} & \multicolumn{1}{c|}{0.8047}    & \multicolumn{1}{c|}{0.8171}        & \multicolumn{1}{c|}{0.8221}    \\ 

\multicolumn{1}{|c|}{DenseNet}            & \multicolumn{1}{c|}{\cellcolor[HTML]{EFEFEF}0.8418}     & \multicolumn{1}{c|}{{\color{blue}0.7986}} & \multicolumn{1}{c|}{0.7866} & \multicolumn{1}{c|}{0.7802} & \multicolumn{1}{c|}{0.7808} & \multicolumn{1}{c|}{0.7789} & \multicolumn{1}{c|}{{\color{red}0.8008}} & \multicolumn{1}{c|}{0.7966}     &\multicolumn{1}{c|}{0.7984} & \multicolumn{1}{c|}{0.7668}  & \multicolumn{1}{c|}{0.7961} & \multicolumn{1}{c|}{0.7781}    & \multicolumn{1}{c|}{0.7829}        & \multicolumn{1}{c|}{0.7878}    \\ 

\multicolumn{1}{|c|}{ViT}                    & \multicolumn{1}{c|}{\cellcolor[HTML]{EFEFEF}0.9163}     & \multicolumn{1}{c|}{{\color{red}0.9041}} & \multicolumn{1}{c|}{0.8923} & \multicolumn{1}{c|}{0.8857} & \multicolumn{1}{c|}{0.8893} & \multicolumn{1}{c|}{0.8881} & \multicolumn{1}{c|}{0.8871} & \multicolumn{1}{c|}{0.8888}     & \multicolumn{1}{c|}{0.8942} & \multicolumn{1}{c|}{0.8772}  & \multicolumn{1}{c|}{{\color{blue}0.9039}} & \multicolumn{1}{c|}{0.8755}    & \multicolumn{1}{c|}{0.8807}        & \multicolumn{1}{c|}{0.8909}    \\ \hline
\end{tabular}}
\label{table_Caltech}
\end{table*}
\begin{table*}[]
\centering
\caption{The top table is a quantitative comparison of overall accuracy for image classification on the proposed Dark Class dataset. The bottom table is a quantitative comparison of mean Average Precision (mAP, \%) for object detection on the ExDark dataset \cite{Exdark}. }
\resizebox{\textwidth}{!}{
\begin{tabular}{|ccccccccccclll|}
\hline
\multicolumn{1}{|c|}{Dataset}         & \multicolumn{13}{c|}{Dark Class}                                                                                                                                                                                                                                                                                                                                                                \\ \hline
\multicolumn{1}{|c|}{Preprocess}      & \multicolumn{1}{c|}{\cellcolor[HTML]{EFEFEF}NULL}   & \multicolumn{1}{c|}{Dong}   & \multicolumn{1}{c|}{EnGAN}  & \multicolumn{1}{c|}{JED}    & \multicolumn{1}{c|}{KinD}   & \multicolumn{1}{c|}{LIME}   & \multicolumn{1}{c|}{RetinexNet} & \multicolumn{1}{c|}{SCI} & \multicolumn{1}{c|}{UTV}    & \multicolumn{1}{c|}{Zero}                        & \multicolumn{1}{c|}{SNR}  & \multicolumn{1}{c|}{PairLIE}  & \multicolumn{1}{c|}{SMG} \\ \hline

\multicolumn{1}{|c|}{VGG}           & \multicolumn{1}{c|}{\cellcolor[HTML]{EFEFEF}0.8144} & \multicolumn{1}{c|}{0.7798} & \multicolumn{1}{c|}{0.7798} & \multicolumn{1}{c|}{0.8095} & \multicolumn{1}{c|}{0.8032} & \multicolumn{1}{c|}{0.7934} & \multicolumn{1}{c|}{0.7782}     & \multicolumn{1}{c|}{0.7903} & \multicolumn{1}{c|}{0.8042} & \multicolumn{1}{c|}{0.7866} & \multicolumn{1}{l|}{{\color{blue}0.8183}}    & \multicolumn{1}{c|}{0.7895}        & {\color{red}0.8196}
 \\ 

\multicolumn{1}{|c|}{ResNet}        & \multicolumn{1}{c|}{\cellcolor[HTML]{EFEFEF}0.7862} & \multicolumn{1}{c|}{0.7907} & \multicolumn{1}{c|}{0.7862} & \multicolumn{1}{c|}{0.7911} & \multicolumn{1}{c|}{0.7903} & \multicolumn{1}{c|}{0.7903} & \multicolumn{1}{c|}{0.7874}     & \multicolumn{1}{c|}{0.7705} & \multicolumn{1}{c|}{{\color{blue}0.8068}} & \multicolumn{1}{c|}{0.7979} & \multicolumn{1}{c|}{{\color{red}0.8171}}    & \multicolumn{1}{c|}{0.7939}        & 0.7958                         \\ 
\multicolumn{1}{|c|}{DenseNet}     & \multicolumn{1}{c|}{\cellcolor[HTML]{EFEFEF}0.8176} & \multicolumn{1}{c|}{0.7791} & \multicolumn{1}{c|}{0.7787} & \multicolumn{1}{c|}{0.7975} & \multicolumn{1}{c|}{0.7834} & \multicolumn{1}{c|}{0.7859} & \multicolumn{1}{c|}{0.7771}     & \multicolumn{1}{c|}{0.7758} & \multicolumn{1}{c|}{0.7954} & \multicolumn{1}{c|}{0.7967} & \multicolumn{1}{c|}{{\color{red}0.8224}}    & \multicolumn{1}{c|}{0.7967}        & {\color{blue}0.8002}                         \\ 
\multicolumn{1}{|c|}{ViT}             & \multicolumn{1}{c|}{\cellcolor[HTML]{EFEFEF}0.8063} & \multicolumn{1}{c|}{0.8135} & \multicolumn{1}{c|}{0.7979} & \multicolumn{1}{c|}{0.8099} & \multicolumn{1}{c|}{0.7978} & \multicolumn{1}{c|}{{\color{blue}0.8242}} & \multicolumn{1}{c|}{0.7831}     & \multicolumn{1}{c|}{0.8047} & \multicolumn{1}{c|}{0.8087} & \multicolumn{1}{c|}{{\color{red}0.8299}} & \multicolumn{1}{c|}{0.8138}    & \multicolumn{1}{c|}{0.7996}        &        0.7982                  \\ \hline \hline
\multicolumn{1}{|c|}{Dataset}         & \multicolumn{13}{c|}{ExDark}                                                                                                                                                                                                                                                                                                                                                                       \\ \hline
\multicolumn{1}{|c|}{Preprocess}   & \multicolumn{1}{c|}{\cellcolor[HTML]{EFEFEF}NULL}   & \multicolumn{1}{c|}{Dong}   & \multicolumn{1}{c|}{EnGAN}  & \multicolumn{1}{c|}{JED}    & \multicolumn{1}{c|}{KinD}   & \multicolumn{1}{c|}{LIME}   & \multicolumn{1}{c|}{RetinexNet} & \multicolumn{1}{c|}{SCI}  & \multicolumn{1}{c|}{UTV}    & \multicolumn{1}{c|}{Zero}   & \multicolumn{1}{c|}{SNR} & \multicolumn{1}{c|}{PairLIE} & \multicolumn{1}{c|}{SMG} \\ \hline
\multicolumn{1}{|c|}{RetinaNet}       & \multicolumn{1}{c|}{\cellcolor[HTML]{EFEFEF}61.44}  & \multicolumn{1}{c|}{61.01}  & \multicolumn{1}{c|}{60.75}  & \multicolumn{1}{c|}{59.72}  & \multicolumn{1}{c|}{61.71}  & \multicolumn{1}{c|}{61.88}  & \multicolumn{1}{c|}{60.23}      & \multicolumn{1}{c|}{{\color{blue}62.75}}  & \multicolumn{1}{c|}{62.07}  & \multicolumn{1}{c|}{62.32}  & \multicolumn{1}{c|}{61.39}    & \multicolumn{1}{c|}{{\color{red}63.17}}        & \multicolumn{1}{c|}{62.04}    \\ 
\multicolumn{1}{|c|}{EfficientDet} & \multicolumn{1}{c|}{\cellcolor[HTML]{EFEFEF}60.68}  & \multicolumn{1}{c|}{59.88}  & \multicolumn{1}{c|}{57.45}  & \multicolumn{1}{c|}{59.09}  & \multicolumn{1}{c|}{57.34}  & \multicolumn{1}{c|}{{\color{blue}60.55}}  & \multicolumn{1}{c|}{56.69}      & \multicolumn{1}{c|}{{\color{red}60.78}}  & \multicolumn{1}{c|}{57.93}  & \multicolumn{1}{c|}{60.16}  & \multicolumn{1}{c|}{59.14}    & \multicolumn{1}{c|}{59.47}        & \multicolumn{1}{c|}{58.50}    \\ 
\multicolumn{1}{|c|}{YOLOv7}         & \multicolumn{1}{c|}{\cellcolor[HTML]{EFEFEF}76.09}  & \multicolumn{1}{c|}{71.67}  & \multicolumn{1}{c|}{73.16}  & \multicolumn{1}{c|}{74.23}  & \multicolumn{1}{c|}{72.54}  & \multicolumn{1}{c|}{{\color{blue}74.73}}  & \multicolumn{1}{c|}{70.22}      & \multicolumn{1}{c|}{{\color{red}75.85}}  & \multicolumn{1}{c|}{73.32}  & \multicolumn{1}{c|}{73.44}  & \multicolumn{1}{c|}{73.03}    & \multicolumn{1}{c|}{73.41}        &        \multicolumn{1}{c|}{72.51}                  \\ \hline
\end{tabular}}
\label{table_Dark_Ex}
\end{table*}

\subsection{Effect of LLIE on Classification and Object Detection}
In our study, we utilized a range of low-light datasets, including Caltech256-LL, Dark Class, and ExDark, to train models for image classification and object detection. To assess the impact of light enhancement, we also trained these models using light-enhanced versions of the same datasets. This dual approach allowed for a comprehensive evaluation of model performance across standard and enhanced lighting conditions.
Our findings, presented in Fig. \ref{fig-line} and Tables \ref{table_Caltech}-\ref{table_Dark_Ex}, reveal that the performance of image classification and object detection algorithms is comparable whether using low-light or light-enhanced images. 
In particular, Transformer-based classifiers show a slight degradation in their classification performance.
Furthermore, the features extracted by ResNet50 on the Dark Class dataset and its light-enhanced ones are similar, as illustrated in Fig. \ref{fig6}.

These observations suggest that when a network achieves exceptional feature extraction capabilities (e.g., Transformer), its performance degradation in low-light scenes is minor. This capability stems from the network can learn more discriminant information from input images for their tasks, leading to impressive classification/detection results. Conversely, the impact of LLIE techniques on improving the performance of high-level tasks is either marginal or negative, regardless of the feature extraction ability of the networks. 
This outcome is primarily because LLIE methods focus on improving human visual perception rather than restoring semantic and structural information. 
Consequently, the network learns poorer quality image features and processes less effective image information, resulting in undermining the overall performance.

\begin{figure}[]
\centering
\includegraphics[width=8.25 cm]{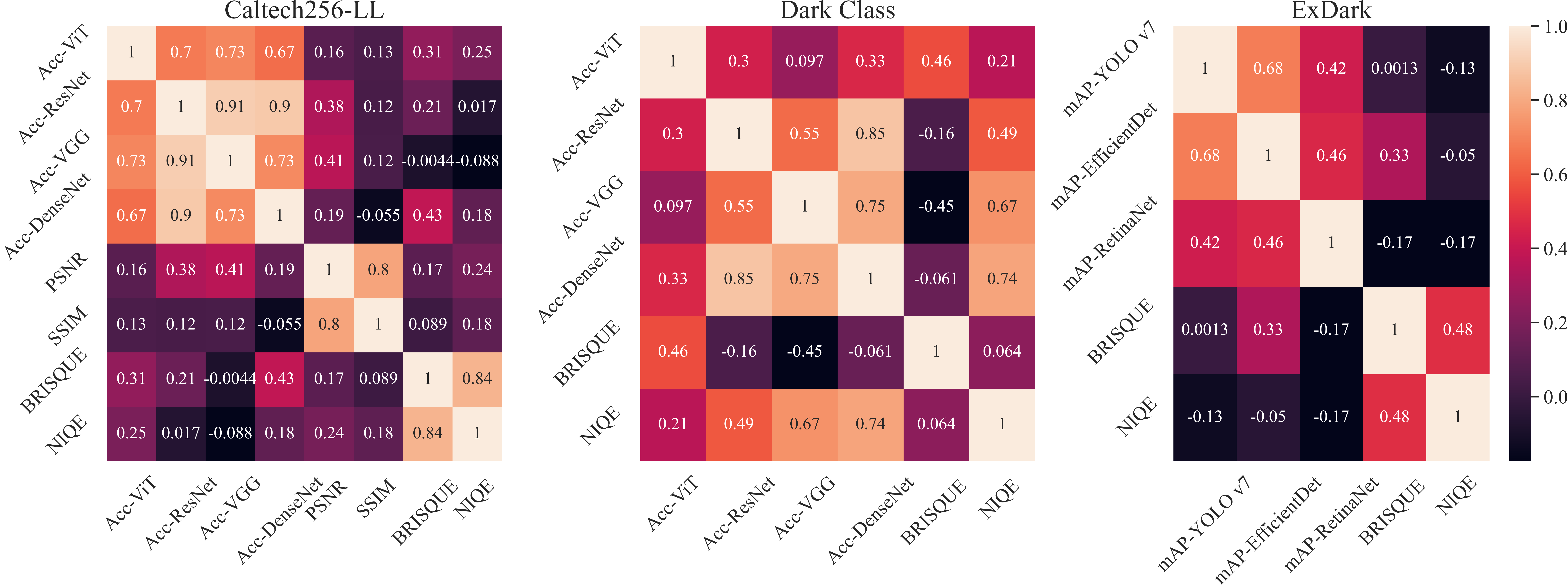}
\caption{The correlation analysis based on Pearson coefficient between image evaluation indices, overall accuracy, and mAP in the results of image classification and object detection.}
\label{fig5}
\vspace{-3mm}
\end{figure}

\begin{figure*}[]
\centering
\includegraphics[width=16 cm]{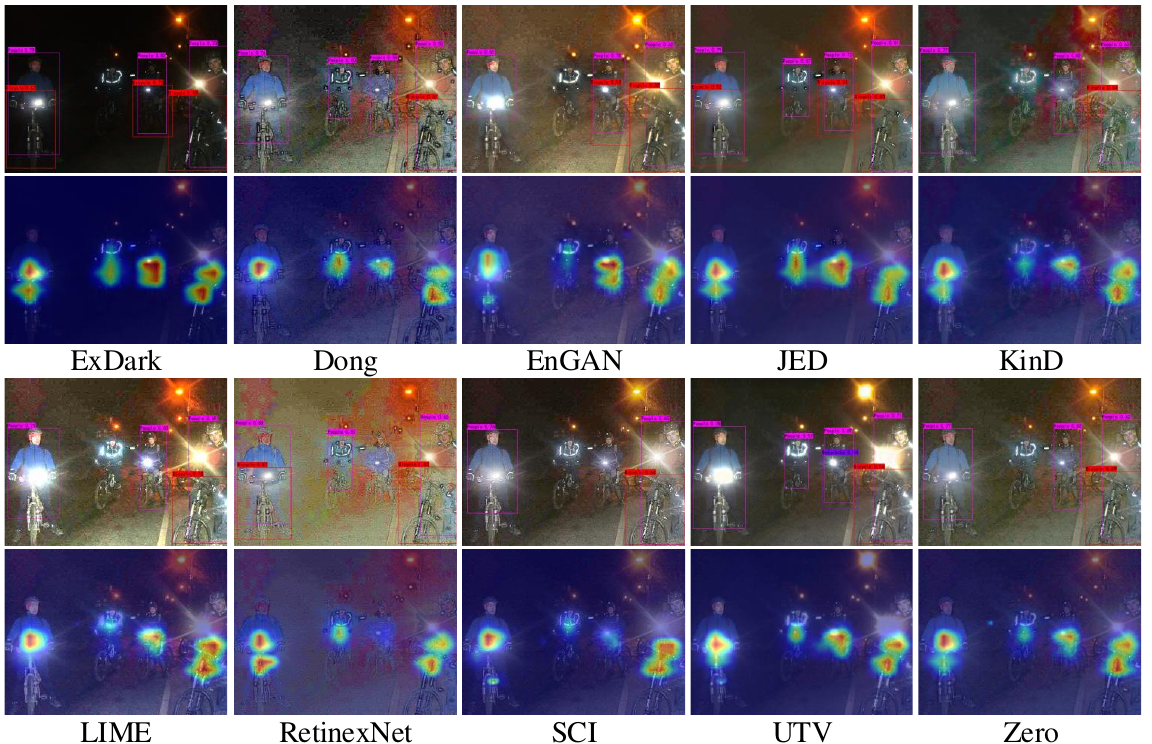}
\caption{Comparisons on object detection. }
\label{fig_obj_compare}
\end{figure*}

\subsection{Discussion}
Building on our extensive experiments and analysis, Fig. \ref{fig5} and Fig. \ref{fig_obj_compare} provide a key insight: High scores in visual perception metrics attained by LLIE methods do not means better performance in high-level tasks. This understanding leads us to engage in a discussion about formulating an efficient LLIE algorithm {aimed to improve} the performance of high-level tasks.

\textit{1) More Information.} To augment the semantic content of images enhanced by LLIE methods, integrating multimodal information presents a promising approach. For instance, textual information can be incorporated into the enhancement using models like CLIP \cite{CLIP}. This integration of textual context with visual data can significantly enrich the semantic {information} of the enhanced images, potentially leading to more meaningful and effective results for high-level tasks.
\textit{2) Robust Learning.} The design of the loss functions should prioritize guiding the LLIE network to learn semantic information and produce images with rich semantic content. For example, we design a loss function to push the network to learn light-invariant features. This loss can be performed by a discriminative loss that is used to determine whether the features extracted by the network are from low or normal-light images. \textit{3) Appropriate Indicators.} {To evaluate image quality, we can use metrics that measure how much an enhanced image improves performance in image classification and object detection. These include indicators like the increase in image classification accuracy and the improvement in object detection's mAP.}


\section{Conclusion}
This study empirically investigates the effectiveness of various LLIE methods in terms of high-level vision tasks such as image classification and object detection. The results show that available LLIE methods do not significantly improve the performance of these tasks and may even reduce it. This suggests that although LLIE methods produce clearer images that are more visually pleasing to humans, they do not improve the semantic and structural information that is crucial for these tasks. These findings are important for computer vision applications that rely on large-scale image datasets, such as surveillance and autonomous vehicles, which require not just visually pleasing images but also images that improve the performance of high-level vision tasks. Therefore, it is important for future research to focus on developing LLIE methods that not only enhance the visibility of low-light images but also improve their performance in high-level vision tasks.

\bibliographystyle{IEEEtran}
\bibliography{main.bib}

\end{document}